\documentclass[letterpaper, 10 pt, conference]{template/IEEEtran}  

\IEEEoverridecommandlockouts                              
\usepackage{graphics} 
\usepackage{graphicx}
\usepackage[tight,footnotesize]{subfigure}
\usepackage{caption}
\usepackage{epsfig} 
\usepackage{mathptmx,cuted} 
\usepackage{times} 
\usepackage{amsmath} 
\usepackage{amssymb}  
\usepackage{hyphenat}
\usepackage[linesnumbered,ruled]{algorithm2e}
\usepackage{color}
\usepackage{amsthm}
\usepackage[utf8]{inputenc}
\usepackage[english]{babel}

\newtheorem*{theorem*}{Theorem}
\newtheorem{theorem}{Theorem}
\newtheorem{example}{Example}
\usepackage[tight,footnotesize]{subfigure}
\long\def\invis#1{}
\newcommand{\C}{\mathcal{C}}
\newcommand{\Cfree}{\C_\textrm{free}}
\newcommand{\Cobs}{\C_\textrm{obs}}
\newcommand{\Csub}{\C_\textrm{sub}}
\newcommand{\qinit}{q_{\rm init}}
\newcommand{\qgoal}{q_{\rm goal}}
\newcommand{\R}{\mathbb{R}}

\def\floatsep{4.0pt}
\def\textfloatsep{0.0pt}
\def\dblfloatsep{4.0pt}
\def\dbltextfloatsep{4.0pt}
\title{\LARGE \bf Analysis of Motion Planning by Sampling in Subspaces of Progressively Increasing Dimension}

\author{Marios P. Xanthidis$^1$, Joel M. Esposito$^2$, Ioannis Rekleitis$^1$, and Jason M. O'Kane$^1$
\thanks{$^1$Marios P. Xanthidis, Ioannis Rekleitis, and Jason M. O'Kane are with the Computer Science and Engineering Department, University of South Carolina,
United States, {\tt\small mariosx@email.sc.edu, [yiannisr, jokane]@cse.sc.edu}%

$^2$Joel M. Esposito is with the Weapons and Systems Engineering Department, United States Naval Academy, {\tt\small esposito@usna.edu}
}
}
\begin{document}

\maketitle
\thispagestyle{empty}
\pagestyle{empty}

\begin{abstract}
Despite the performance advantages of modern sampling-based motion planners, solving high dimensional planning problems in near real-time remains a challenge. Applications include hyper-redundant manipulators, snake-like and humanoid robots. Based on the intuition that many of these problem instances do not require the robots to exercise every degree of freedom independently, we introduce an enhancement to popular sampling-based planning algorithms aimed at circumventing the exponential dependence on dimensionality. We propose beginning the search in a {\em lower dimensional subspace} of the configuration space in the hopes that a simple solution will be found quickly. After a certain number of samples are generated, if no solution is found, we increase the dimension of the search subspace by one and continue sampling in the higher dimensional subspace. In the worst case, the search subspace expands to include the full configuration space --- making the completeness properties identical to the underlying sampling-based planer. Our experiments comparing the enhanced and traditional version of RRT, RRT-Connect, and BidirectionalT-RRT on both a planar hyper-redundant manipulator and the Baxter humanoid robot indicate that a solution is typically found much faster using this approach and the run time appears to be less sensitive to the dimension of the full configuration space. We explore important implementation issues in the sampling process and discuss its limitations.

%
 
\end{abstract}


\section{Introduction} 
\label{sec_intro}

It is well known that the general motion planning problem is
PSPACE-complete~\cite{reif1985complexity} and that the runtime of even the best known exact
algorithm is exponential in the dimension of the configuration space~\cite{canny1988complexity}.  While sampling-based motion planning algorithms
such as rapidly exploring random trees (RRTs)~\cite{lavalle1998rapidly} and probabilistic
roadmaps (PRMs)~\cite{kavraki1996probabilistic} are able to avoid explicit reconstruction of
the free configuration space Canny's algorithm~\cite{canny1988complexity} relies upon, they cannot avoid the curse of dimensionality.
In particular, Esposito's Conditional Density Growth Model (CDGM) for
RRTs~\cite{esposito2013conditional}, predicts that the expected number of samples
required for an RRT to explore a certain volume fraction of a space remains
exponential in the dimension.  For this reason, even though modern sampling-based
planners have exhibited dramatic improvements over the original RRT and PRM
techniques, computing real time solutions for systems with 10 or more degrees
of freedom, such as hyper-redundant manipulators, snake-like robots or
humanoids, remains a significant challenge.


This study focuses on this path planning problem in high-dimensional
configuration spaces. The intuition arises from three observations:
\begin{itemize}
 \item For many systems, it is rare for all of the kinematic abilities to be
 needed for a specific task~\cite{RekleitisMed2016}.

 \item Several studies that attempt to reduce the dimensionality in motion planning in various
 ways have already been
 done~\cite{van2005prioritized,wagner2015subdimensional,vernaza2011efficient,gipson2013resolution,gochev2011path,yoshida2005humanoid,kim2015efficient,wells2015adaptive}.

 \item As shown by {\c{S}}ucan and Kavraki~\cite{alexandru2009performance}, even
 random projections of the configuration space can provide good estimates for
 its coverage.
\end{itemize}

\begin{figure}[t]
\centering
\includegraphics[width=0.40\textwidth, clip=true, trim=0.0in 0.0in 0.0in 0.0in]{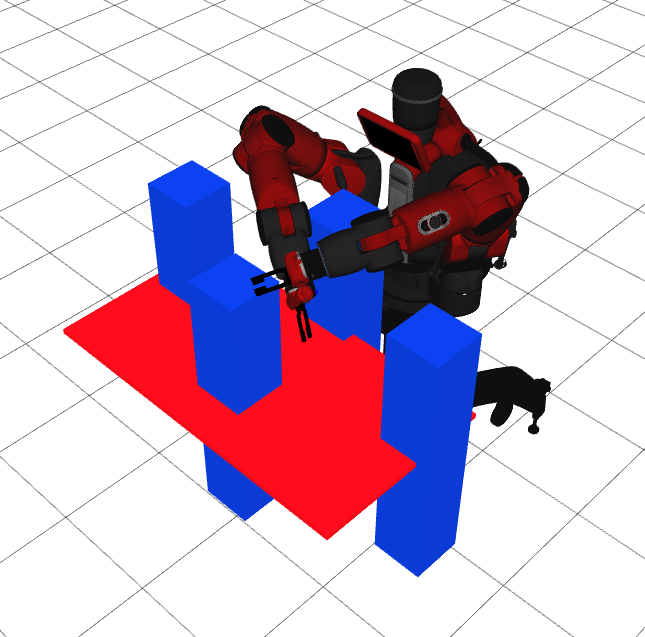}
\caption{The simulated Baxter humanoid robot used in the experiments.}
\label{fig:baxter}
\end{figure}

Based on this intuition, we explore an algorithmic enhancement aimed at reducing the expected
runtime of several common sampling-based planners. We propose beginning the search in a {\em lower
dimensional subspace} of $\cal{C}$ in the hopes that a simple solution will be found quickly. After a
certain number of samples are generated, if no solution is found, we increase the dimension of the
search subspace by 1 and continue sampling in the larger subspace. We repeat this process until a
solution is found. In the worst case, the search subspace expands to include the full dimensional
configuration space --- making completeness properties identical to the
underlying sampling-based planer. In many cases our experiments indicate
that a solution is found much faster using this approach and the run time
appears to be less sensitive to the full dimension of the configuration space.

%
An important property of these subspaces is that a solution entirely lying
inside these subspaces should be feasible, in the absence of obstacles, so the
initial and goal configurations must lie inside every subspace.  The proposed
method, by construction, generates subspaces that satisfy this constraint.

To evaluate this approach, we modified three well established planners, RRT~\cite{lavalle1998rapidly}, RRT-Connect~\cite{kuffner2000rrt}, and BidirectionalT-RRT~\cite{jaillet2008transition} to produce RRT$^+$, RRT$^+$-Connect, and BidirectionalT-RRT$^+$, with the $^+$ symbol indicating that the planners are enhanced using the idea described above. All three planners were compared to the originals and
to KPIECE~\cite{csucan2009kinodynamic} and STRIDE~\cite{gipson2013resolution}.  These planners were tested on a planar hyper-redundant arm and on a simulated Baxter humanoid robot, shown in Figure~\ref{fig:baxter}, through OMPL~\cite{sucan2012open} and the MoveIt! framework~\cite{chitta2012moveit}.
The results show that even a seemingly straightforward method for defining such
subspaces can still have superior planning time performance for many problems
compared to state-of-the-art planners.

The remainder of this paper is structured as follows. Section~\ref{sec:background} reviews other approaches on planning for high-dimensional systems. Section III provides technical details of the enhancement, considering important issues such as how the search subspaces is selected, how to generate the samples,
and when to expand the search dimension. Section IV presents our experience enhancing three well
established planners RRT~\cite{lavalle1998rapidly}, RRT-Connect~\cite{kuffner2000rrt}, and BidirectionalT-RRT~\cite{jaillet2008transition}. Finally, Section V concludes with a discussion of future work.

\section{Related work}
\label{sec:background}

The problem of motion planning has been proven to be PSPACE-hard~\cite{reif1985complexity}. During the late 1990's, sampling-based methods were introduced and shown to be capable of solving challenging motion planning problems, but without guarantees of finding the solution in finite time~\cite{lavalle2006planning}. The two most prominent representatives of those algorithms are probabilistic roadmaps (PRMs) by Kavraki \emph{et al.}~\cite{kavraki1996probabilistic} that are useful for multiple queries in a known, generally stable environment, and RRTs by LaValle~\cite{lavalle1998rapidly}, that are more suitable for single query applications. Two other variations of RRTs, RRT-Connect by Kuffner and LaValle~\cite{kuffner2000rrt} that extends the tree more aggressively in each iteration, and T-RRT by Jaillet \emph{et al.}~\cite{jaillet2008transition} that plans efficiently in costmaps, were used in this study.

Although the performance of these techniques can be affected substantially by the degrees of freedom, some studies have used them successfully for high-dimensional configuration spaces. A method that uses PRMs and finds collision-free paths for hyper-redundant arms was presented by Park \emph{et al.}~\cite{park2011collision}. Other studies use RRTs for motion planning of redundant manipulators, such as the work of Bertram \emph{et al.}~\cite{bertram2006integrated}, which solves the inverse kinematics in a novel way.  Weghe \emph{et al.}~\cite{weghe2007randomized} apply RRTs to redundant manipulators without the need to solve the inverse kinematics of the system. A study by Qian and Rahmani~\cite{qian2013path} combines RRTs and inverse kinematics in a hybrid algorithm in a way that drives the expansion of RRTs by the Jacobian pseudo-inverse. 

Additionally, some works use RRTs for mobile manipulators. Vannoy \emph{et al.}~\cite{vannoy2008real} propose an efficient and flexible algorithm for operating in dynamic environments.  The work of Berenson \emph{et al.}~\cite{berenson2008optimization} provides an application of their technique to a 10-DoF mobile manipulator.

For multi-robot systems, many sampling-based algorithms have been proposed.  The study of van den Berg and Overmars~\cite{van2005prioritized} uses a PRM and presents a prioritized technique for motion planning of multiple robots. Other studies use RRT-based algorithms such as the study by Carpin and Pagello~\cite{carpin2002parallel} which introduced the idea of having multiple parallel RRTs for multi-robot systems. The work of Wagner~\cite{wagner2015subdimensional} plans for every robot individually and, if needed, coordinates the motion in higher dimensional spaces. Other studies propose efficient solutions by using a single RRT~\cite{otani2009applying, solovey2015finding}. 

Furthermore, motion planning for humanoid robots with sampling-based algorithms has been explored; Kuffner \emph{et al.} presented algorithms for motion planning on humanoid robots with both the use of PRMs~\cite{kuffner2002dynamically} and RRTs~\cite{kuffner2005motion}.  Other studies, such as the work of Liu \emph{et al.}~\cite{liu2012hierarchical}, use RRTs for solving the stepping problem for humanoid robots. 

\invis{But while there are many different efficient algorithms for different purposes based on sampling, as it has been already described, there is very little work on proposing a general method for dealing with big configuration spaces.}  The work of Vernaza and Lee tried to extract structural symmetries in order to reduce the dimensionality of the problem~\cite{vernaza2011efficient} by also providing near-optimal solutions. This technique is more time efficient than the traditional RRT, mostly for very high-dimensional configuration spaces. \invis{and only for known environments where the cost function is more or less stable.} Yershova \emph{et al.}~\cite{yershova2005dynamic} proposed an approach to focus sampling in the most relevant regions.

\begin{figure*}[thpb]
 \begin{center}
  \leavevmode
   \begin{tabular}{ccc}
     \subfigure[]{\includegraphics[width=0.25\textwidth]{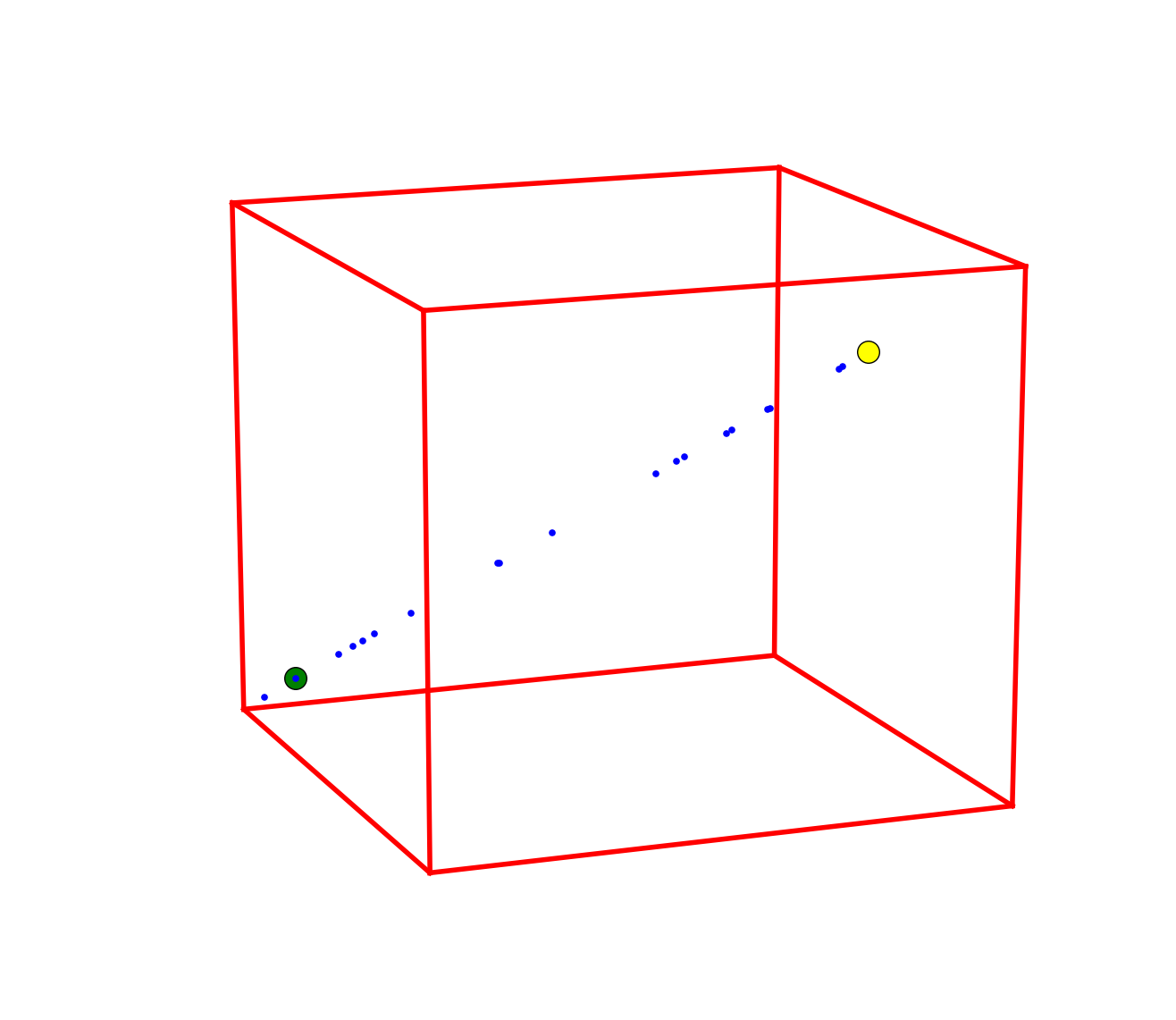}}&
     \subfigure[]{\includegraphics[width=0.25\textwidth]{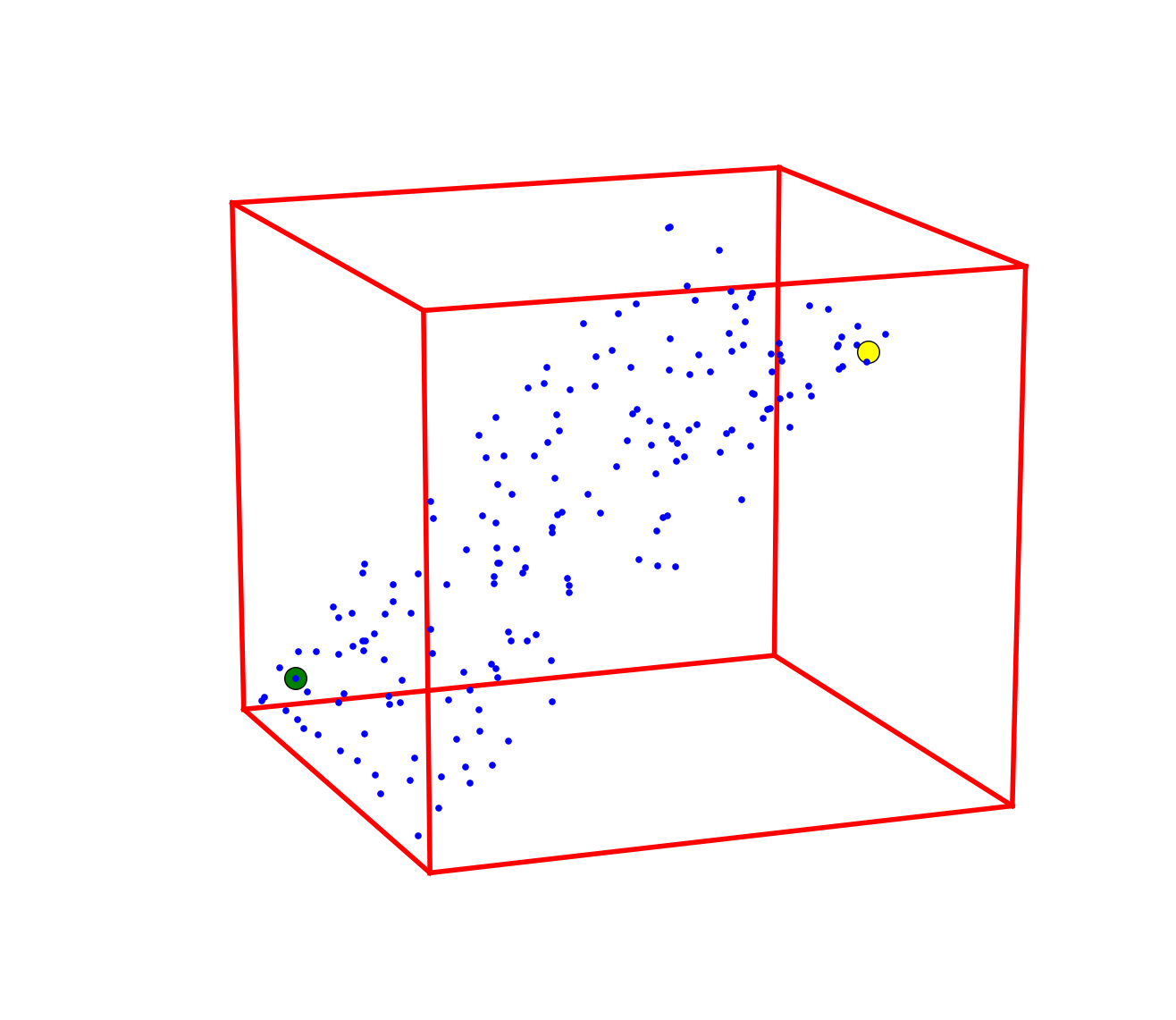}}&
     \subfigure[]{\includegraphics[width=0.25\textwidth]{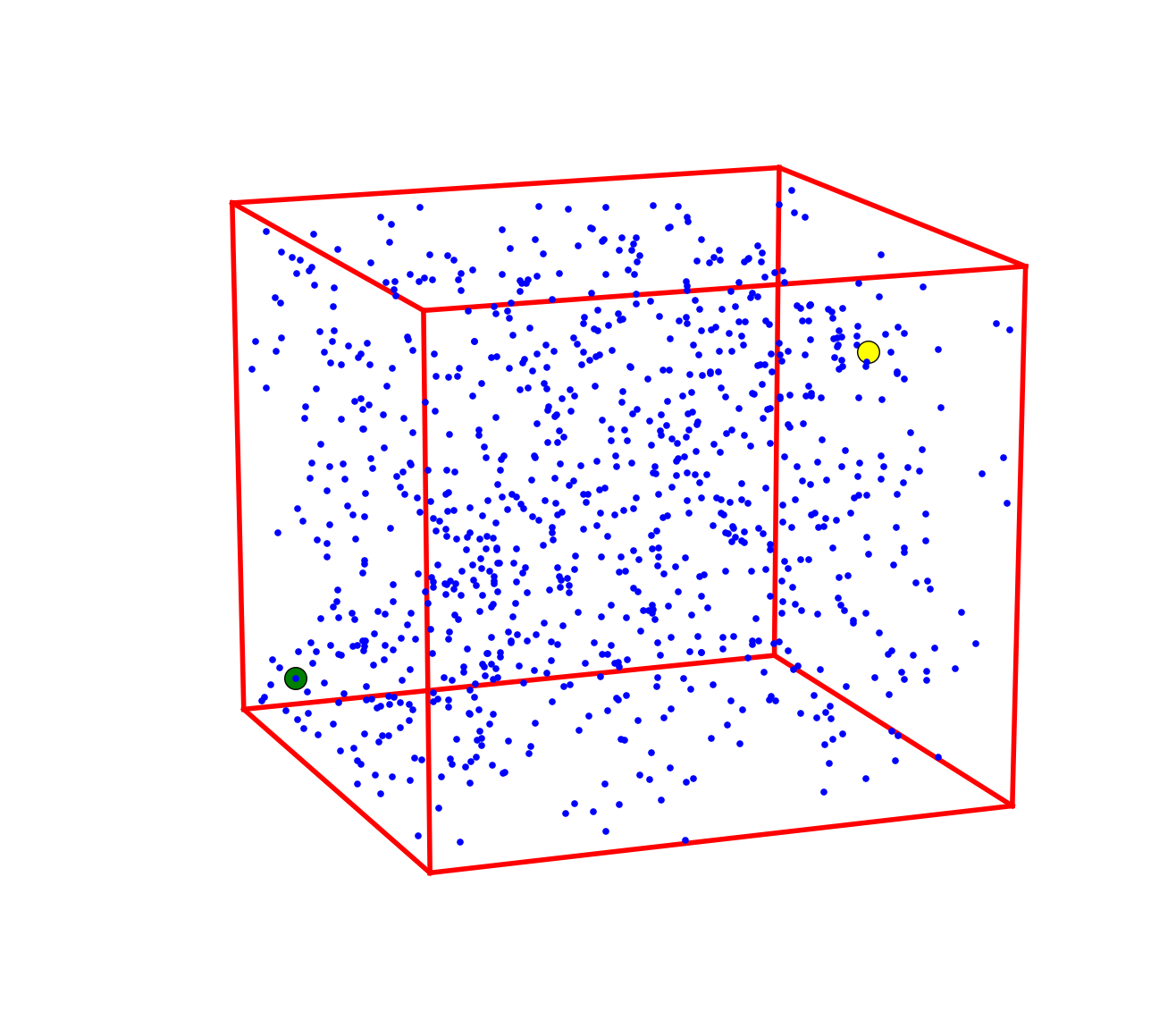}}
    \end{tabular}
  \end{center}
\caption{Sampling in one, two and three dimensions in $\C$. Red lines indicate the boundaries of $\C$, yellow indicate $q_{init}$, and green indicate $q_{goal}$.}
\label{fig:samples}
\end{figure*}

Finally, other approaches attempt to deal with the dimensionality in different ways. Gipson \emph{et
al.}~developed STRIDE~\cite{gipson2013resolution} which samples non-uniformly
with a bias toward unexplored areas of the configuration space. KPIECE~\cite{csucan2009kinodynamic} by {\c{S}}ucan and Kavraki uses random 2D and 3D projections to estimate the coverage of $\C$. \invis{where the density of samples is lower,
provided a fast planner for high-dimensional configuration spaces.}
Additionally, Gochev \emph{et al.}~\cite{gochev2011path} proposed a motion
planner that decreases the dimensionality by recreating a configuration space
with locally adaptive dimensionality. Yoshida's work~\cite{yoshida2005humanoid}
tries to sample in ways that exploit the redundancy of a humanoid system.  \invis{by modeling the system
with different ways that each one forms a subspace in the configuration space
and then with the use of a specific sample generator for every different
configuration model and the adaptive changes of the configurations of the
system the planner provides fast solutions.} Kim \emph{et al.}~\cite{kim2015efficient} present an RRT-based algorithm for articulated robots
that reduces the dimensionality of the problem by projecting each sample into
subspaces that are defined by a metric. Shkolnik and Tedrake ~\cite{shkolnik2009path} plan for highly redundant manipulators
in the dimensional task space with the use of Jacobian transpose and Voronoi bias. Wells and Plaku~\cite{wells2015adaptive} reduce the dimensionality for 2-D hyper-redundant manipulators by modeling the end-effector as a single mobile robot, and the other links as trailers that are getting pulled. Lastly,
the work of Bayazit \emph{et al.}~\cite{bayazit2005iterative} where a PRM was used to plan in subspaces of the configuration space,
creates paths that solve an easier problem than the original and then iteratively optimize the solution until the solution becomes valid. 

This paper introduces the idea of iteratively searching in lower-dimensional subspaces, and emphasizes the potential of using such approach for efficient motion planning on arbitrary hyper-redundant systems. The approach operates regardless of if there is an articulated robot, a humanoid, or a heterogeneous multi-robot system, in a sense that additional limitations are not introduced in the original planners. Unlike previous works, the approach tries to find paths that are not only confined entirely to a subspace but also in a subspace in which a solution can exist, since the initial and goal configurations are part of the subspace. Contrary to the work of Bayazit \emph{et al.}~\cite{bayazit2005iterative} the approach searches in subspaces that are strictly lower-dimensional. Intuitively, instead of finding results faster by simplifying the problem, our approach tries to find plans fast by simplifying the model of the system by applying virtual constraints, and then by iteratively relaxing them.\invis{so the tree tries to expand only in a way that satisfies these constraints.}

\section{Methodology}
\label{sec:Method}

\subsection{Problem Statement}
Let $\C$ denote a configuration space with $n$ degrees of freedom, partitioned
into free space $\Cfree$ and obstacle space $\Cobs$ with $\C= \Cfree \cup
\Cobs$.
The obstacle space $\Cobs$ is not explicitly represented, but instead can be
queried using collision checks on single configurations or short path segments.
Given initial and goal configurations $\qinit, \qgoal \in \Cfree$, we
would like to find a continuous path within $\mathcal{C}_{\rm free}$ from
$\qinit$ to $\qgoal$.

\newcommand{\ssmin}{^{\rm (min)}}  
\newcommand{\ssmax}{^{\rm (max)}}  
\newcommand{\cint}[1]{\left[c_{#1}\ssmin, c_{#1}\ssmax\right]}  

For purposes of sampling, we assume that each degree of freedom in $\C$ is
parameterized as an interval subset of $\R$, 
so that
  \begin{equation}
    \C = \cint{1} \times\dots\times \cint{n} \subseteq \R^n. \label{eq:C}
  \end{equation}
Note that we treat $\C$ as Euclidean only in the context of sampling; other
operations such as distance calculations and the generation of local path segments
utilize identifications on the boundary of $\C$ as appropriate for the
topology.

\subsection{Planning in Subspaces}
The proposed method is based on searching for a solution in lower
dimensional subspaces of $\C$, in hopes that such a path might be found faster
than expanding the tree in all dimensions. 
The underlying idea is to exploit the redundancy of each system for each
problem.  To achieve this, the algorithm starts searching in the unique linear 1-dimensional
subspace of $\C$ that contains $\qinit$ and $\qgoal$.  If this search fails, the planner expands its search
subspace by one dimension.  This process continues iteratively until the
planner finds a path, or until it searches in all of $\C$. In each subsearch the tree structure created in lower dimensions is kept and
expanded in subsequent stages.

\begin{algorithm}[t]
  \caption{RRT$^+$}\label{RRT+}
  \SetKwInOut{Input}{Input}
  \SetKwInOut{Output}{Output}
  \DontPrintSemicolon
  \SetAlgoLined
  \Input{A configuration space $\C$, an initial configuration $\qinit$, and a goal configuration $\qgoal$.}
  \Output{RRT graph $G$}

  $G$.init($q_{init}$)\;
  $\Csub \gets $ 1-d subspace of $\C$, through $q_{init}$ and $q_{goal}$\; \label{lineA}

  \While{True}{
    $q_{\rm rand} \gets$ sample drawn from $\Csub$ \; \label{lineC}
    $q_{\rm near} \gets $NearestVertex$(q_{\rm rand},G)$\;
    $q_{\rm new} \gets $NewConf$(q_{\rm near}, q_{\rm rand})$\;
    $G$.AddVertex($q_{\rm new}$)\;
    $G$.AddEdge($q_{\rm near}$, $q_{\rm new}$)\;
    \If{done searching $\Csub$\label{lineD}}{
      \uIf{$\dim(\Csub) < \dim(\C)$}{
        Expand $\Csub$ by one dimension. \label{lineB}
      }
      \Else{
        \Return $G$
      }
    }
  }
\end{algorithm}

Algorithm~\ref{RRT+} summarizes the general approach, as applied to RRT.
The planner starts optimistically by searching in one dimension, along the line
passing through $\qinit$ and $\qgoal$.  If this search fails to find a path---a
certainty, unless there are no obstacles between $\qinit$ and $\qgoal$---the
search expands to a planar subspace that includes $\qinit$ and $\qgoal$, then
to a 3D flat\footnote{We use the term \emph{flat} to refer to a
  subset of $\R^n$ congruent to some lower-dimensional Euclidean space.}, and so on until, in the worst case, the algorithm
eventually searches all of $\C$; see Figure~\ref{fig:samples}. \invis{It should be noted that, although for simplicity the method described in this study uses linear constraints since it only depends on the constraints between the DoF, it can potentially be generalized into defining complex lower-dimensional manifolds if appropriate constraints are applied. }

\invis{The pseudocode in Algorithm~\ref{RRT+} shows a straightforward,
single-directional RRT$^+$, analogous to the standard RRT algorithm.}
\invis{Note, however, the idea readily applies to most tree-based motion planners,
since the primary difference is in how the samples are generated.
The experiments in Section~\ref{Experim}, for example, describe RRT$^+$ planners
based on the well-known goal biased and bidirectional RRT algorithms.}

The description in Algorithm~\ref{RRT+} leaves three important elements
unspecified.
First, the algorithm needs a method for selecting and representing the subspace
$\Csub$ (Lines~\ref{lineA} and \ref{lineB}).
Second, a method is required for sampling from this subspace (Line~\ref{lineC}).
Third, the conditions that must be met before moving to
the next subsearch must be defined (Line~\ref{lineD}).
The choices explored in this study are described in the next sections.
\invis{Particular choices for each of these elements can be based on knowledge about
the specific types of problems being solved; this freedom is why we refer to
RRT$^+$ as a family of algorithms.  Sections~\ref{sec:subspace},
\ref{sec:sampling}, and \ref{sec:termination} describe possible
instantiations of the basic framework.  Section~\ref{sec:pc} establishes conditions
under which the resulting planner is probabilistically complete.}


\subsection{Representing and Sampling from Subspaces}\label{sec:subspace}

The central idea is to search for solutions in subspaces of progressively
higher dimensions.  The primary constraint on these subspaces is that they must
contain both $\qinit$ and $\qgoal$. Subspaces that violate this constraint
cannot, of course, contain a path connecting $\qinit$ to $\qgoal$.

In general, the algorithm's selections for $\Csub$ should ideally be directed
by the likelihood that a solution will exist fully within $\Csub$.

\newcommand{\Pcon}{P_{\rm con}}
%
  The choice that we investigate in this paper---one that trades some generality for simplicity--- is a
  prioritized release of the degrees of freedom.
  Given a set $\Pcon \subseteq \{1, \ldots, n \}$ of DoFs to be
  constrained, we can form $\Csub$ by constraining the Dofs in $\Pcon$ to 
  form a line passing from $\qinit$ and $\qgoal$ and allowing the remaining DoF to vary freely.

Next, the algorithm requires a technique for drawing samples from $\Csub$. The sampling happens in a very efficient linear time method \invis{to produce samples in the intersection of hyperplanes passing
through $\qinit$ and $\qgoal$ is the prioritized sampling as described in Example~3. }
that initially generates a sample within $\C$
along the line between $\qinit$ and $\qgoal$ by selecting a random scalar $r$
and applying:
  \begin {equation} \label{get_sample}
    q_{\rm rand}^{(i)}= (\qgoal^{(i)}-\qinit^{(i)})r+\qinit^{(i)}
  \end{equation}
The algorithm then modifies $q_{\rm rand}$ by inserting, for each DoF not in
$\Pcon$, a different random value within the range for that dimension; see Algorithm~\ref{sampler}.

To ensure that the original sample along the line $L$ between $\qinit$ and $\qgoal$ 
is within $\C$, we compute $r_{min}$ and $r_{max}$ using Algorithm \ref{ComputeBoundaryValues} and select $r$ randomly from the interval $[r_{min}, r_{max}]$. 
The ComputeBoundaryValues function finds the intersections of this line all the different $c_i^{(min)}$ and $c_i^{(max)}$ flats, and defines 
the limits of $r$ from the values of $r$ producing the two intersected points that lie inside the $\C$. The ComputeBoundaryValues function is called only once before the planning loop.

\invis{
To ensure that the original sample in $\C$ along the line between $\qinit$ and $\qgoal$ is within $\C$,

If  $0\leq r \leq 1$ then samples between $q_{init}$ and $q_{goal}$ are
generated, so there is the need to scale properly the $r$ so $ratio_{min}\leq r
\leq ratio_{max}$ where $ratio_{min} \leq 0$, $ratio_{max} \geq 0$ and both
when applied to Equation \ref{get_sample} with $Dimension(P_{con}) =
Dimension(\mathcal{C})$, they give the two different intersection points on the
borders. In the Scale function those values are found by finding the
intersection of all the hyper-planes of the borders of $\mathcal{C}$ with the
line, and keeping the two values that give points in the $\mathcal{C}$. This
function is also used by the RRT$^{+}$ at the initializations of line 2, so it
is a process that happens only once during the planning. Additionally the DoF
that are released are sampled from their entire range defined by the
$\mathcal{C}$ as it is shown in Algorithm \ref{sampler} which is used in line 6 of RRT$^{+}$.}

\begin{algorithm}[t]
\caption{Prioritized Sampler}
\label{sampler}
\SetKwInOut{Input}{Input}
\SetKwInOut{Output}{Output}
\DontPrintSemicolon
\SetAlgoLined
\Input{Initial configuration $\qinit$, goal configuration $\qgoal$, set of constrained DoFs $\Pcon$, boundaries of the random number $r$ $r_{min}$, $r_{max}$.}
\Output{Sample configuration $q$}

$q \gets $ random point in $\C$ along line L from $\qinit$ to $\qgoal$  with $r_{min}$ and $r_{max}$\;
\For{$i \in 1,\ldots,n$}{
	\If{$i \notin \Pcon$}{
	  $q[i] \gets$ Random($0$, $1$)$*(c^{(max)}_i-c^{(min)}_i)+c^{(min)}_i$\;
	}
}

\Return $q$\;
\end{algorithm}

\begin{algorithm}[t]
\caption{ComputeBoundaryValues}
\label{ComputeBoundaryValues}
\SetKwInOut{Input}{Input}
\SetKwInOut{Output}{Output}
\DontPrintSemicolon
\SetAlgoLined
\Input{Initial configuration $q_{init}$, goal configuration $q_{goal}$, dimensionality of configuration space $n$, limits of configuration space $c^{(min)}$, $c^{(max)}$.}
\Output{Minimum value of scalar $r_{min}$, maximum value of scalar $r_{max}$}

$D \gets \qgoal - \qinit$ \;
$L = Dt+q_{init}$\;
\For{\normalfont $i \gets 1$ \textbf{to} $n$}{
\For{\normalfont $c$ \textbf{ in } $\{ c_i^{(min)}, c_i^{(max)} \}$}{
	Find the intersection $p=Dt_p+q_{init}$ of $L$ and $c$\;
	\If {$p \in \C$}{
		Find $t_p$ so $p=Dt_p+q_{init}$\;
		
		\eIf{($t_p\leq 0$)}{
				$r_{min} \gets t_p$\;
			}
			{$r_{max} \gets t_p$\;}
	}
		
}
}
\Return $r_{min}$, $r_{max}$.\;
\end{algorithm}

\invis{
\begin{algorithm}[t]
\caption{Scale}
\label{scale}
\SetKwInOut{Input}{Input}
\SetKwInOut{Output}{Output}
\DontPrintSemicolon
\SetAlgoLined
\Input{Initial configuration $q_{init}$, goal configuration $q_{goal}$, dimensionality of configuration space $n$,limits of configuration space $c^{(min)}$, $c^{(max)}$.}
\Output{Minimum value of scalar $r_{min}$, maximum value of scalar $r_{max}$}

$D \gets \qgoal - \qinit$ \;

\For{\normalfont $i \gets 1$ \textbf{to} $n$}{
\For{\normalfont $c$ \textbf{ in } $\{ c_i^{(min)}, c_i^{(max)} \}$}{
	$w \gets [0, 0, \dots,0]$\;
	$l \gets [0, 0, \dots,0]^T$\;
	$l[i] \gets 1$\;
	$w[i] \gets c$\;
	
	\eIf{($D*l=0$)}{
		\textbf{continue}\;
	}{
		$t \gets \frac{(w-q_{init})*l}{D*l}$\;
		
		\If {($tD+q_{init}$ \textbf{in} $\mathcal{C}$)}{
			\eIf{($t\leq 0$)}{
				$r_{min} \gets t$\;
			}
			{$r_{max} \gets t$\;}
		
		}
	}
}
}
\Return $r_{min}$, $r_{max}$.\;
\end{algorithm}

}

\invis{
While this approach is simple, it loses some generality because only a finite number of flats can be explored, determined by $\qinit$, $\qgoal$, and $\Pcon$. Although the number of the different lower-dimensional projections is finite, practically it is not very limiting, since the number of different projections $L$ for an $n$ DoF system is:
\begin{equation}
L = 2^n -n
\end{equation}
}
The prioritized method provides an efficient and easy way to sample from projections of arbitrary dimensionality, while, as stated before, it provides valuable understanding that serves the purpose of our study.
\invis{By using the formal algebraic description of a hyperplane, every subspace can be formalized as:
\begin{equation}
a_1x_1+a_2x_2+...+a_ix_i+...+a_{n-1}x_{n-1} +a_nx_n = b
\end{equation}
So the subspace in every stage with dimensionality $Dim$ will be:
\begin{equation}
ax+a_1x_1+a_2x_2+...+a_ix_i+...+a_{Dim-1}x_{Dim-1} = b
\end{equation}

So the hyperplane in each iteration of the algorithm can be found with:
\begin{itemize}
\item $a=\sum D[i]$ and $ ratio_{min}\leq x\leq ration_{max}$, for every $i\in P_{con}$
\item $a_i=c^{(max)}_i-c^{(min)}_i$ and $0\leq x_i \leq 1$, for every $i\in P_{ind}$
\item $b=-(Sum_{con}+Sum_{free})$ where $Sum_{con}= \sum {q_{init} [i]}$ for every $i\in P_{con}$ and $Sum_{free}= \sum {C_{min} [i ]}$ for every $i \in P_{ind}$
\end{itemize}}

\subsection{Terminating the subsearches}\label{sec:termination}

The only remaining detail to be discussed is how long the search in each subspace should continue. A set of timeouts \{$t_1, t_2, \ldots, t_n$\} should be generated where $t_i$ corresponds to the amount of time spent in the $i^{\rm th}$ iteration. Ideally, the planner should stop searching in subspaces that seem unlikely to provide a solution. For simplicity in this paper we precompute the timeouts by assuming that the $t_i$ follow a geometric progression. The idea is to exponentially increase the number
of samples in each successive subsearch, acknowledging the need for more
samples in higher dimensions.

The proposed approach uses two different parameters specified by the user. The first parameter $T$ is the timeout for the entire algorithm. The second parameter $\alpha>1$ is a factor describing a constant ratio of the runtime between successive subsearches. The total time $T$ available to the algorithm can be expressed in terms of $\alpha$ and a base time $t_0$:
\begin{equation}
T = \sum_{i=1}^{n} {t_0 \alpha^i}
 \end{equation}
Solving for $t_0$ we obtain:
 \begin{equation}
t_0 = \frac{\alpha-1}{\alpha(\alpha^n-1)}T
\label{T_eq}
 \end{equation}

Using this $t_0$ \invis{in Algorithm~\ref{findsize}}every $t_i$ can be computed as:
\begin{equation}
t_i= \alpha t_{i-1}
\end{equation}

In this study values for $\alpha$ ranging between $1$ and $2$ and $T$ scaling linearly with the dimensionality of the configuration space seem to lead to good performance.
\section{Experiments}
\label{Experim}


For the experiments three new planners were developed using
the OMPL framework~\cite{sucan2012open}.  These planners work by applying the technique to three RRT variants: (1) RRT~\cite{lavalle1998rapidly} with default goal bias of $0.05$,
(2) RRT-Connect~\cite{kuffner2000rrt}, and (3) the bidirectional T-RRT~\cite{jaillet2008transition} by assuming a uniform costmap.
The T-RRT variant is intended to test the effect of our method to a powerful costmap planner.

\subsection{Experiments with a 2D hyper-redundant manipulator}
In order to test the ability of the new planners to adapt to different problems, 
the prioritization was chosen randomly for each run, although an optimization 
might be expected to give results faster. We demonstrate that given enough redundancy even a random prioritization will provide results much faster. A computer with 6th Generation Intel Core i7-6500U Processor (4MB Cache, up to 3.10 GHz) and 16GB of DDR3L 
(1600MHz) RAM was used.

\begin{figure}[t]

   \begin{tabular}{cc}
     \subfigure[]{\fbox{\includegraphics[width=0.197\textwidth]{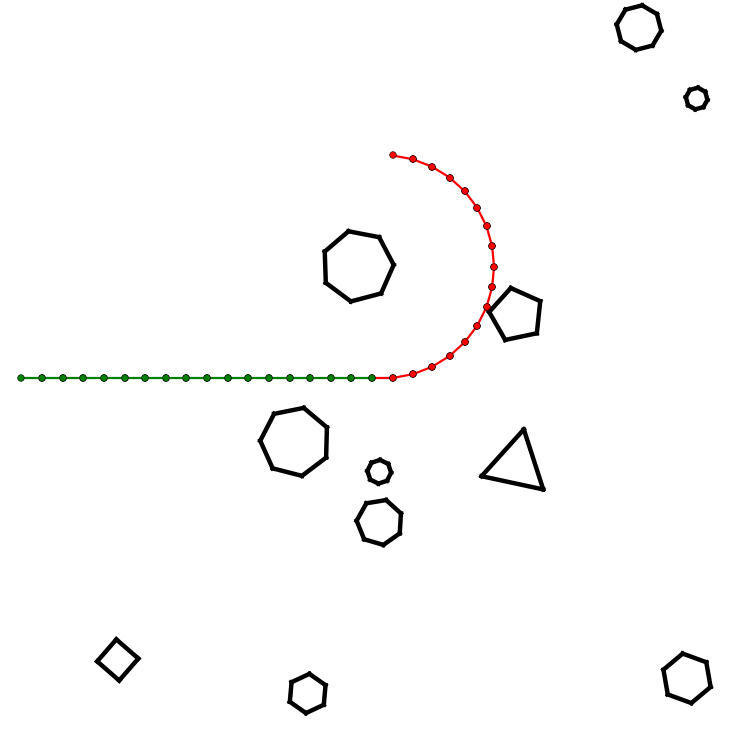}}}&
     \subfigure[]{\fbox{\includegraphics[width=0.205\textwidth]{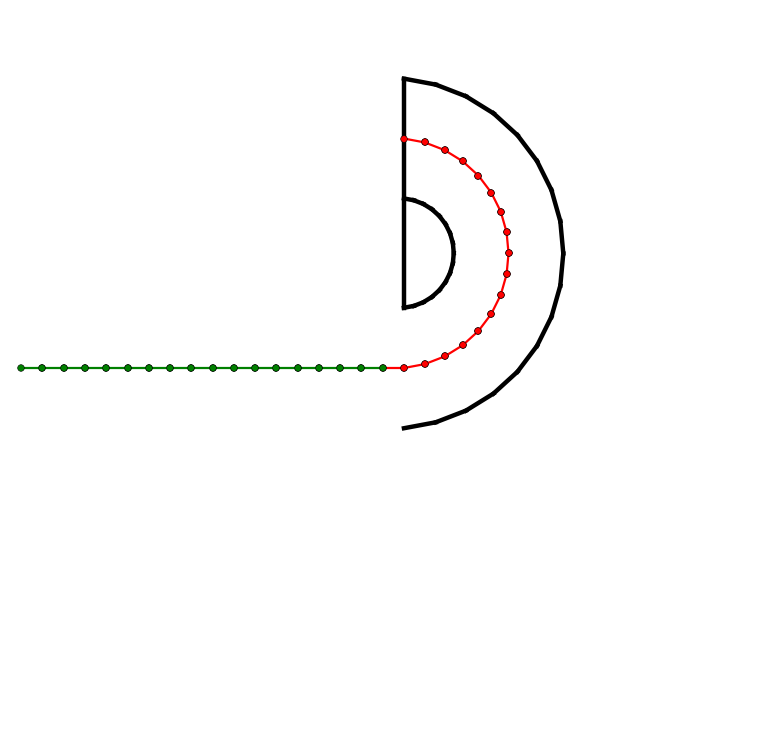}}}\\
     \subfigure[]{\fbox{\includegraphics[width=0.1995\textwidth]{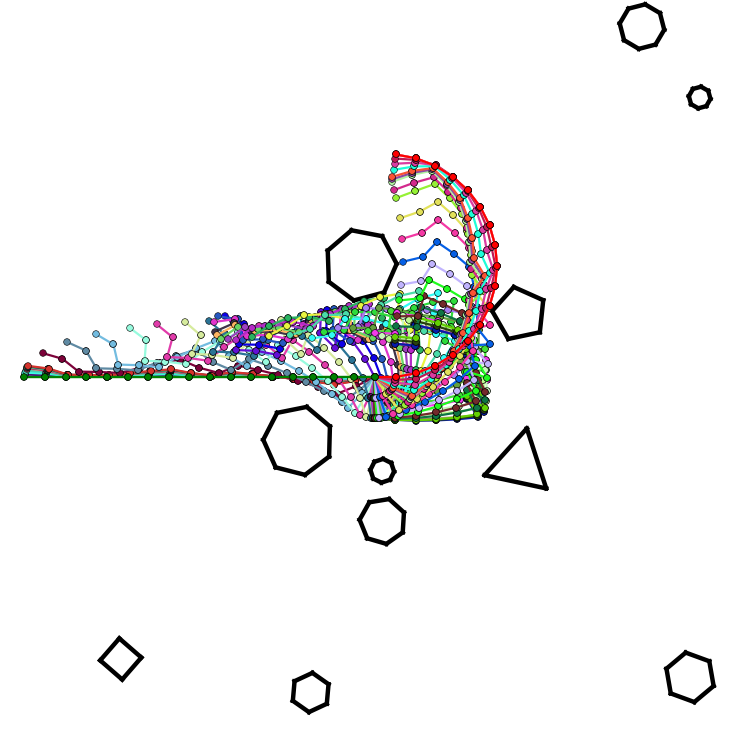}}}&
     \subfigure[]{\fbox{\includegraphics[width=0.204\textwidth]{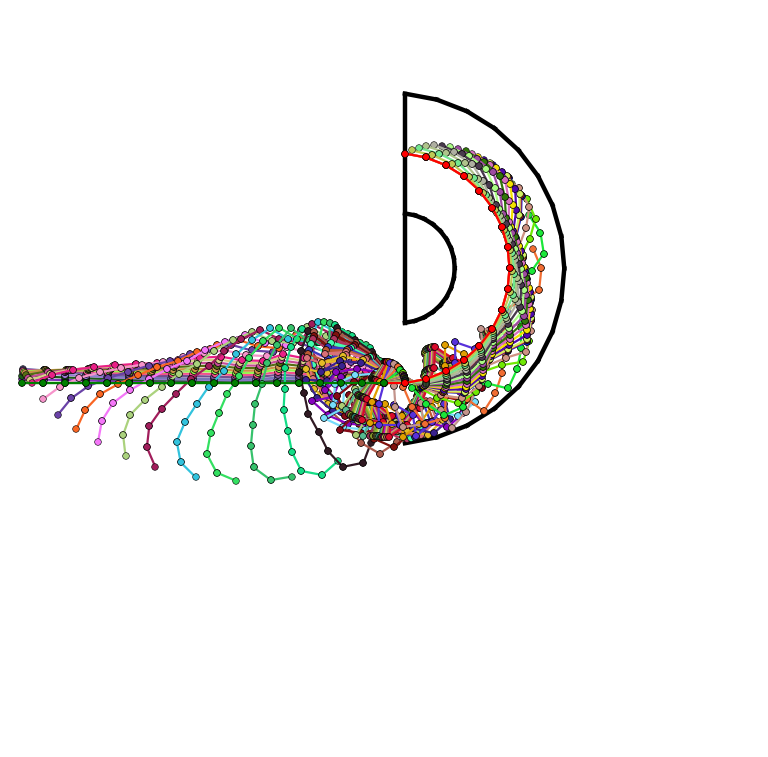}}} 
     \end{tabular}

\caption{The two different environments, called Cluttered Random and Horn, used in \cite{gipson2013resolution}, (a-b) with red indicating $\qinit$ and green $\qgoal$, and two different solutions for those environments (c-d) with RRT$^+$-Connect, respectively.}
\label{fig:res1}
\end{figure}

\invis{\subsection{Experiments with a 2D hyper-redundant manipulator}}
The OMPL's standard example of a 2D hyper-redundant manipulator, created for STRIDE~\cite{gipson2013resolution}, was used in order to test and compare the enhanced planners in a standard way. Two different environments were tested 100~times each for a kinematic chain with varying 12 to 20 degrees of freedom: A Cluttered Random environment and a Horn environment; see Figure~\ref{fig:res1}. The initial and goal configurations were the same as shown in Figure~\ref{fig:res1}, so a qualitative comparison in problems with different redundancies could be done. \invis{Also all the DoF were starting and finishing with the same angle, the motions of the resulted paths, were some DoF were moving freely and the others were forced to rotate by the same angle, will make it more easy to visually study the method.}

\begin{figure}[t]

   \begin{tabular}{c}
     \subfigure[]{\fbox{\includegraphics[width=0.40\textwidth]{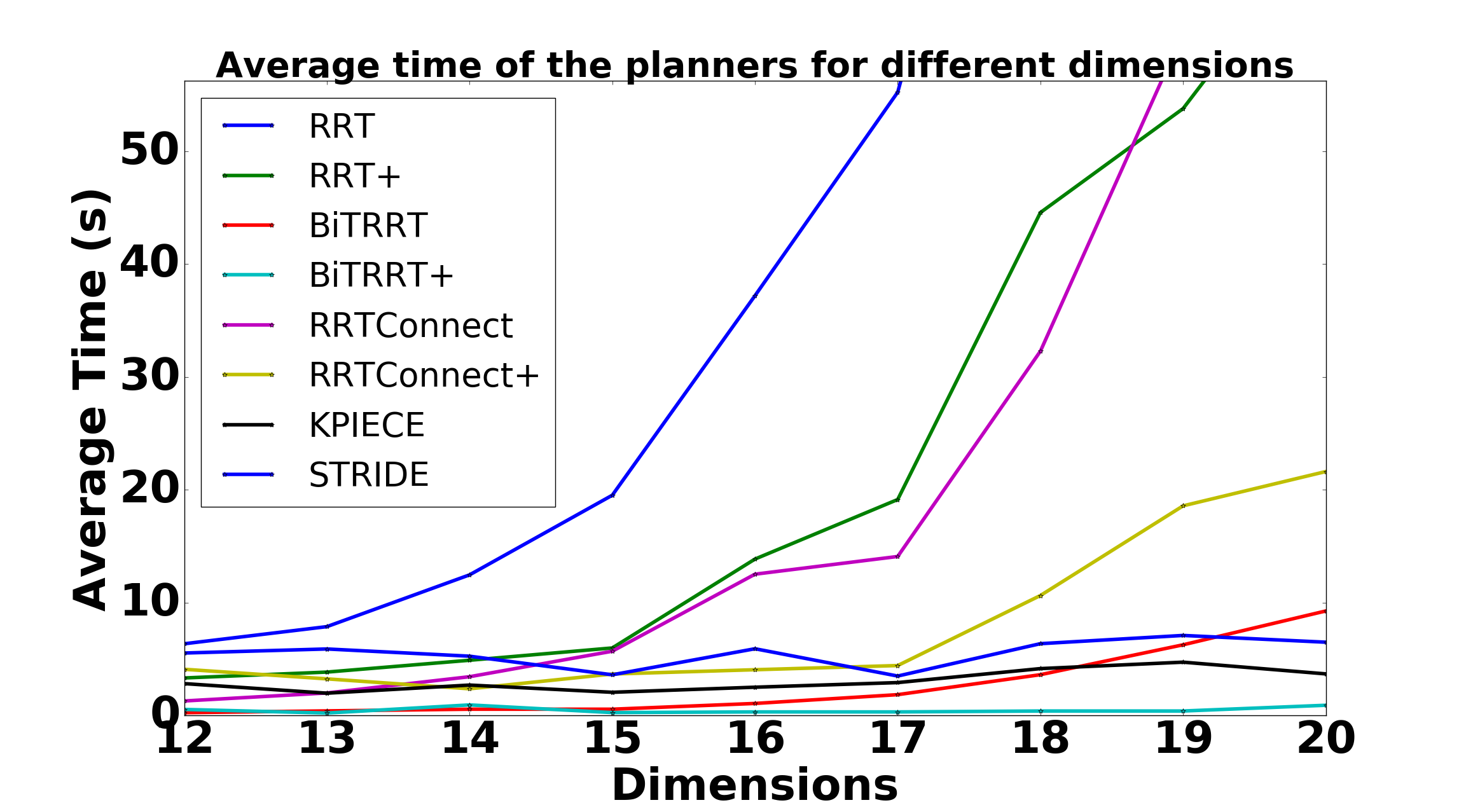}}}\\
     \subfigure[]{\fbox{\includegraphics[width=0.40\textwidth]{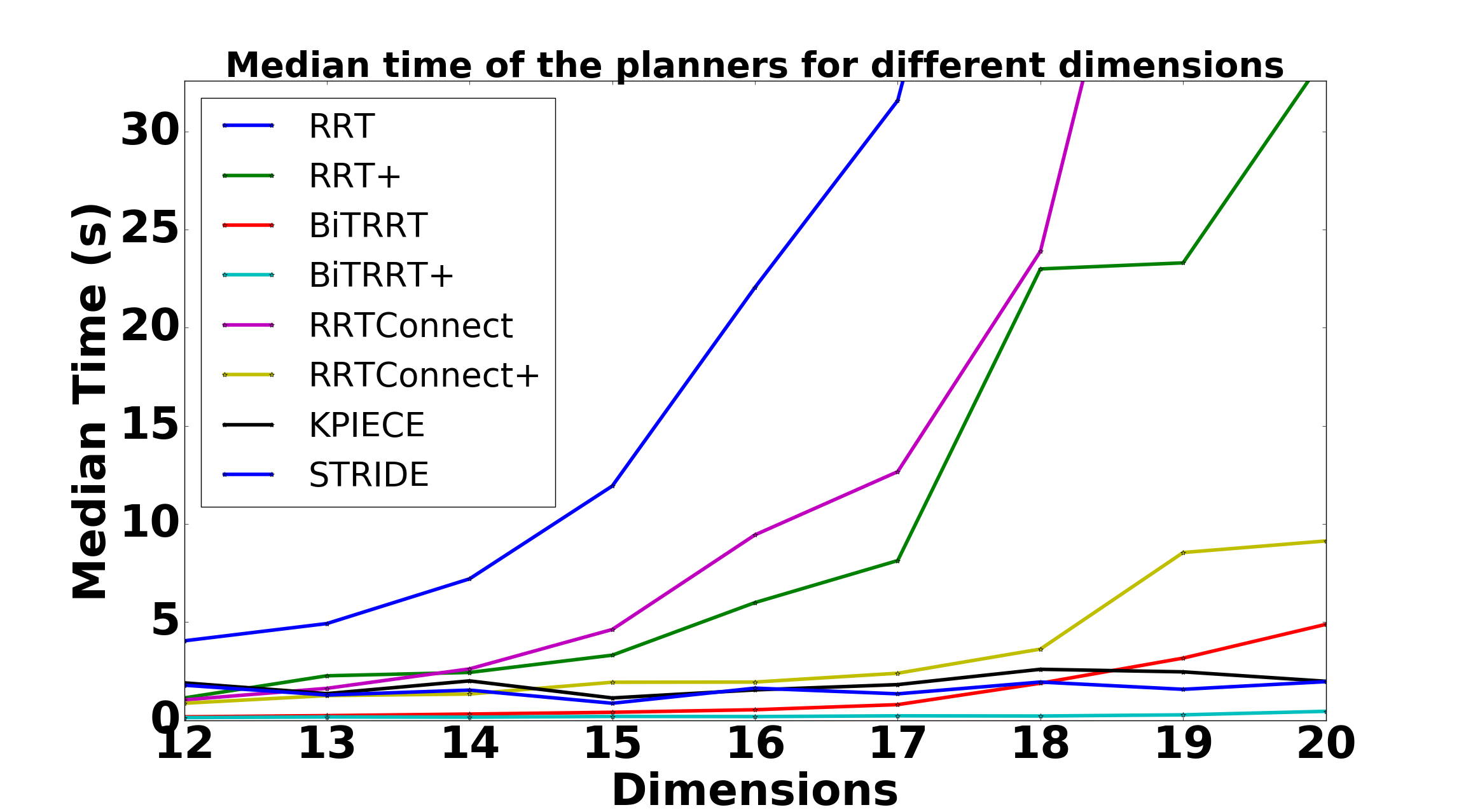}}}
     \end{tabular}

\caption{The average (a) and the median (b) time for the Random Cluttered environment.}
\label{fig:clut}
\end{figure}

\begin{figure}[t]

   \begin{tabular}{c}
     \subfigure[]{\fbox{\includegraphics[width=0.40\textwidth]{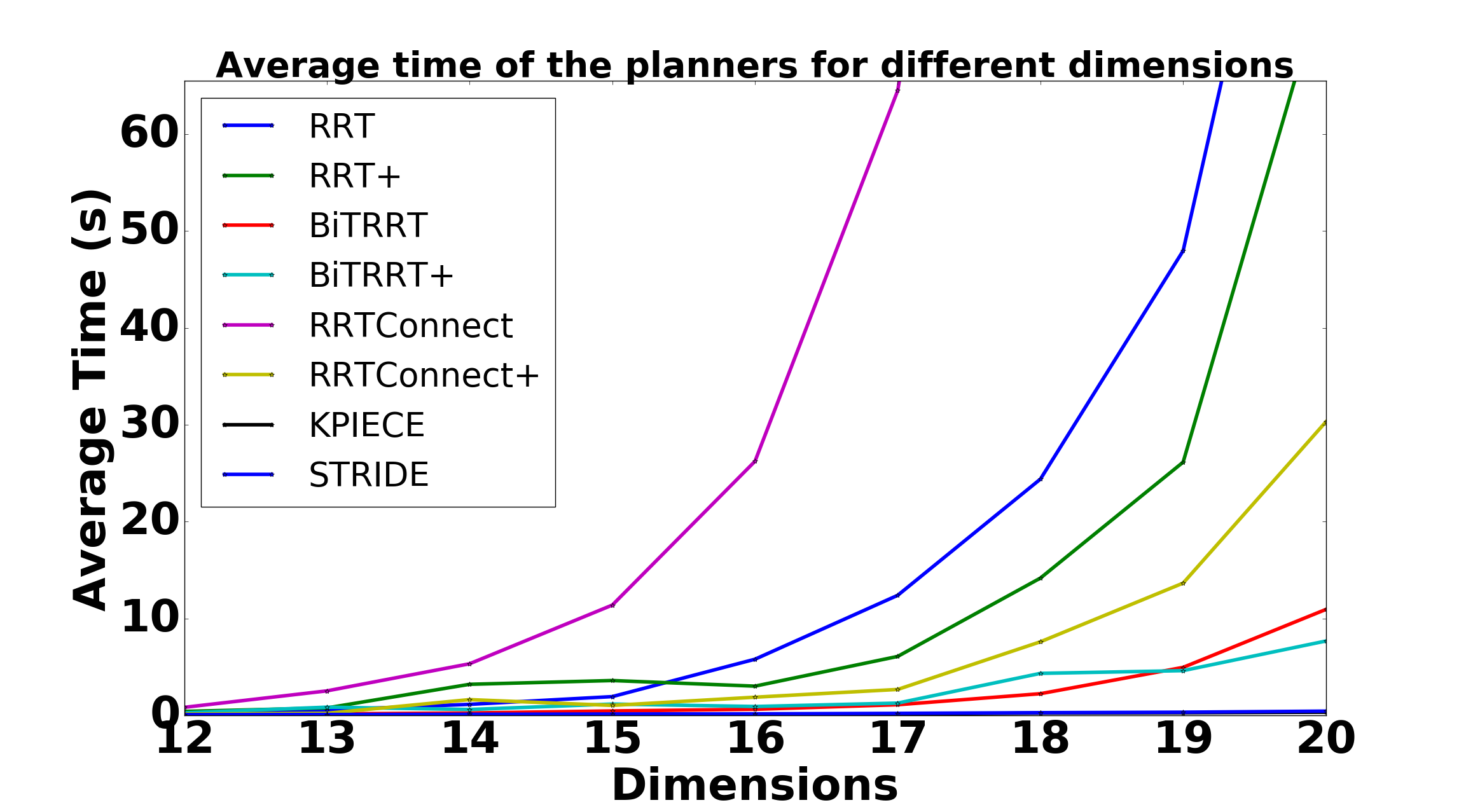}}}\\
     \subfigure[]{\fbox{\includegraphics[width=0.40\textwidth]{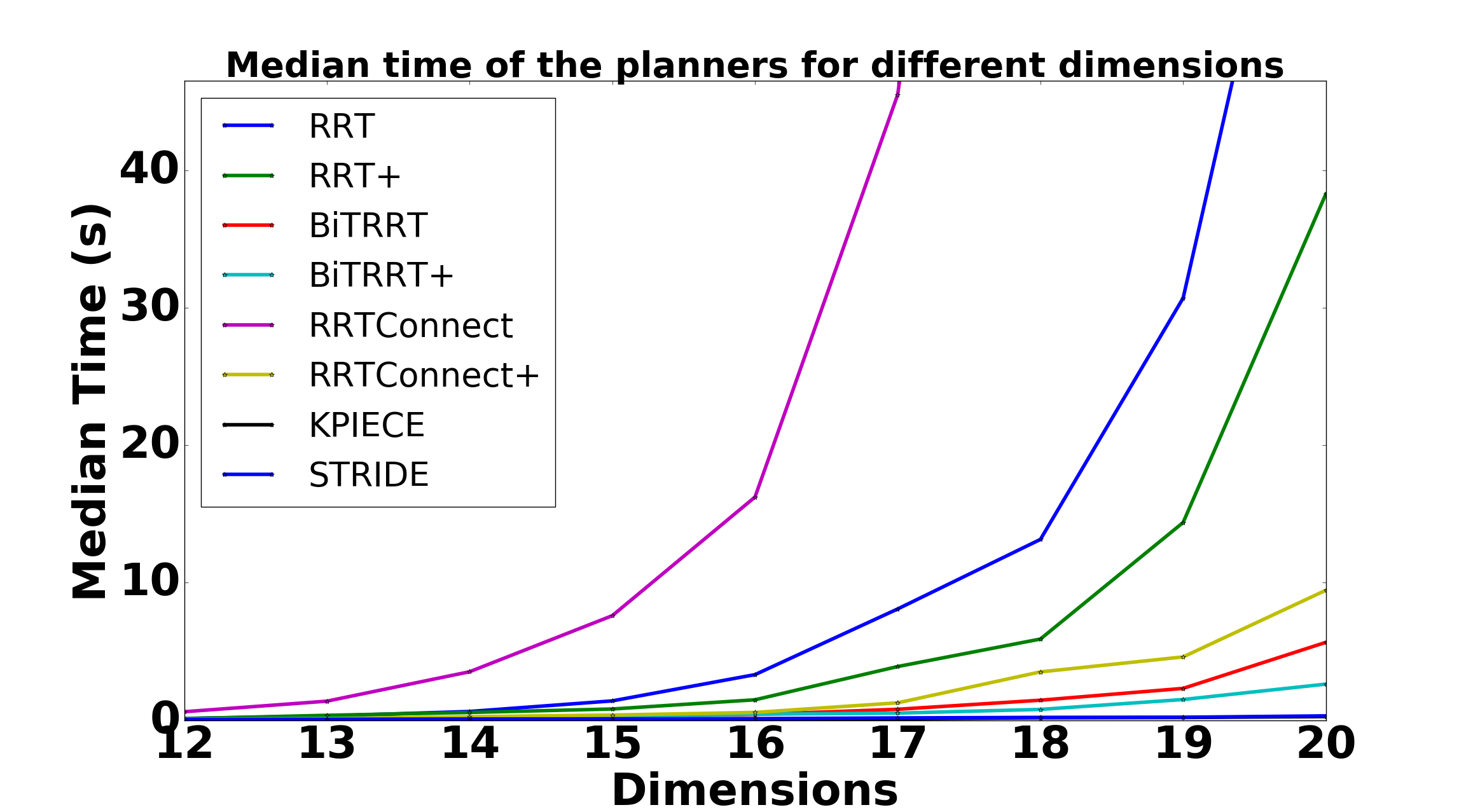}}}
     \end{tabular}

\caption{The average (a) and the median (b) time for the Horn environment.}
\label{fig:horn}
\end{figure}

In all the cases, the enhanced versions of each planner were faster and in most cases significantly faster than the original ones.  The average and the median times for the Random Cluttered environment are presented in Figure~\ref{fig:clut}, and in Figure~\ref{fig:horn} for the Horn environment. Although the plots of the path length are not provided, our planners produced solutions with length --- after the standard simplification method provided by OMPL --- very similar, if not slightly better, than the other planners.

The BiT-RRT$^+$ not only outperformed the BiT-RRT by a wide margin but also outperformed
all the other planners using uniform sampling. Additionally, it provided competitive results to robust planners with biased sampling such as KPIECE~\cite{csucan2009kinodynamic} and STRIDE~\cite{gipson2013resolution}, which as expected perform much better in less redundant environments. 

Interestingly, for each problem, the single fastest solution across all trials
was generated by BiT-RRT$^+$. This suggests that a good choice of
prioritization may give results faster in a consistent way.

\begin{figure}[t]
\centering
\fbox{\includegraphics[width=0.40\textwidth, clip=true]{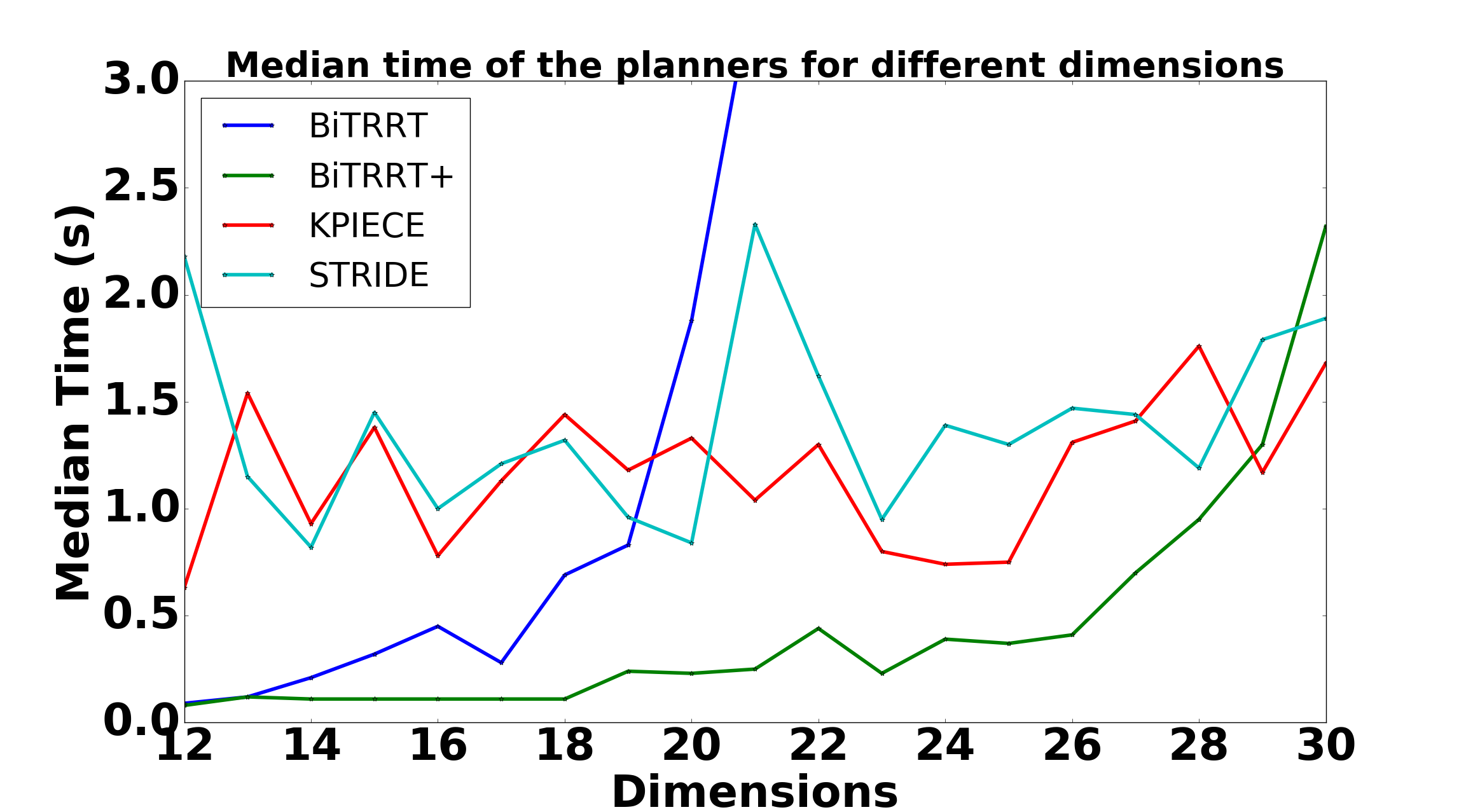}}
\caption{Comparison of the median time for the Cluttered Random environment from 12 to 30 dimensions of BiT-RRT and BiT-RRT$^+$, KPIECE and STRIDE. Interestingly, BiT-RRT$^+$ is more than 200 times faster than BiT-RRT for 30 DoF.}
\label{fig:med_int}
\end{figure}

Additional experiments with the fastest planners were done from 12 to 30 dimensions for the Cluttered environment. The performance of the BiT-RRT$^+$ in Figure \ref{fig:med_int}, it deals much better with the dimensionality. Also the BiT-RRT$^+$ provides the fastest results among all the planners for the most part. Even with random prioritization only after the 29 dimensions due to insufficient prioritizations the BiT-RRT$^+$ performs slower than KPIECE and STRIDE. 

\invis{
\begin{figure}[t]
\centering
\fbox{\includegraphics[width=0.47\textwidth, clip=true]{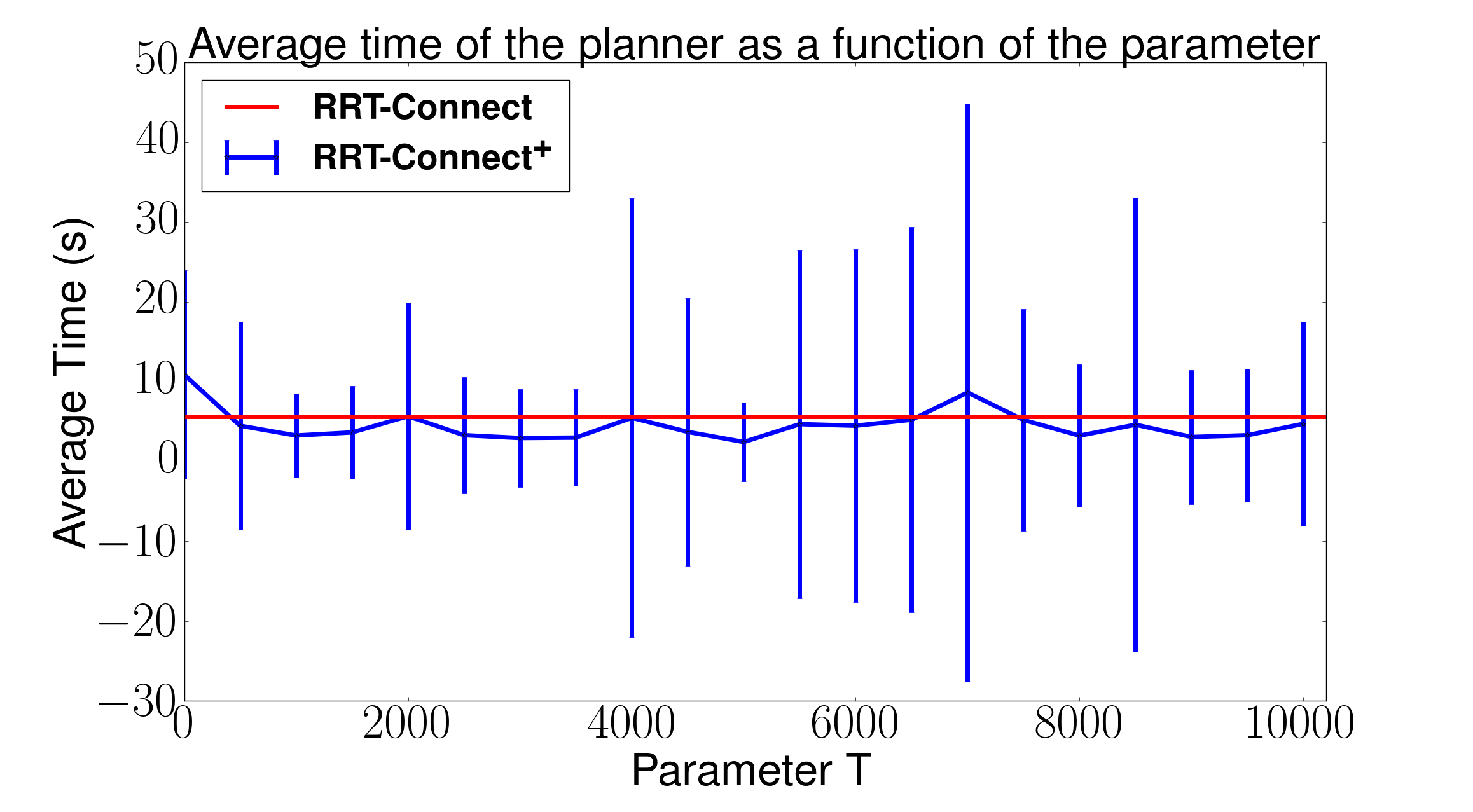}}
\caption{Sensitivity of $T$ from 0 to 10000 with step 500, given $\alpha=1.6$ for 200 runs per value of RRT$^+$-Connect in the Random Cluttered environment for 15 degrees of freedom. The timeout was set to infinity and $T$ was used only for Equation~\ref{T_eq}.}
\label{fig:sens}
\end{figure}
}
\invis{
Lastly, we demonstrate that the parameter $T$ is not sensitive to the performance of the planners for a wide range, Figure~\ref{fig:sens}, although a good tuning may positively affect the efficiency. No obvious pattern is observed on the efficiency of the enhanced planner, proving that mostly insufficient prioritizations affect negatively the plan time. 
}

\subsection{Experiments with Baxter}

Experiments are presented for the Baxter humanoid robot (Figure \ref{fig:baxter}) of 14 degrees of freedom using the OMPL~\cite{sucan2012open} and MoveIt! framework~\cite{chitta2012moveit} with an Intel i7-7700 8-core processor (3.6GHz), and 32 GiB RAM.

Instead of choosing the prioritizations randomly, a generic task independent policy was used to show that even naive policies can eliminate the outliers observed in the other experiments and lead to superior performance. The policy was giving priority to the joints closer to the base.

\begin{figure}[t]

   \begin{tabular}{cc}
     \subfigure[]{\fbox{\includegraphics[width=0.21\textwidth]{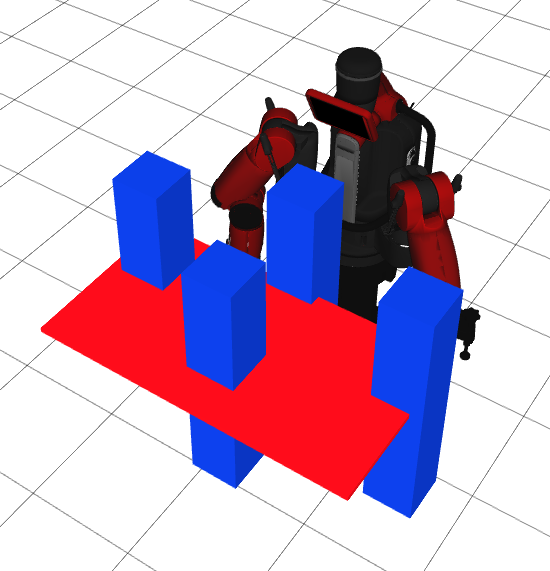}}}&
     \subfigure[]{\fbox{\includegraphics[width=0.21\textwidth]{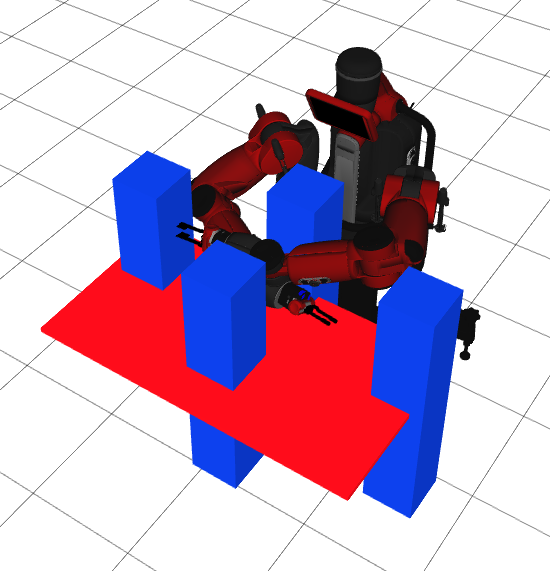}}}
     \end{tabular}

\caption{The initial (a) and goal (b) configuration used in the experiments with the Baxter. The table is indicated with red, and the four pillars are indicated with blue}
\label{fig:bax_init_goal}
\end{figure}

We tested the enhanced planners in a cluttered workspace, consisted by a table and four parallel pillars. As shown in Figure \ref{fig:bax_init_goal}, Baxter starts with the manipulators in the relaxed position below the table and the goal is to reach the configuration that both arms fit between the pillars. Moreover, in order to make the problem more challenging for the enhanced planners, one of the manipulators should end up above the other, reducing even more the redundancy of the problem, forcing our planners to explore subspaces with high dimensionality. The RRT$^+$, RRT$^+$-Connect, BiT-RRT$^+$ were compared with their original versions, with KPIECE and bidirectional version of KPIECE, provided in OMPL, called BiKPIECE for 100 trials. A timeout of 60 seconds was used for each run. 

\begin{figure}[thpb]

   \begin{tabular}{c}
     \subfigure[]{\fbox{\includegraphics[width=0.45\textwidth]{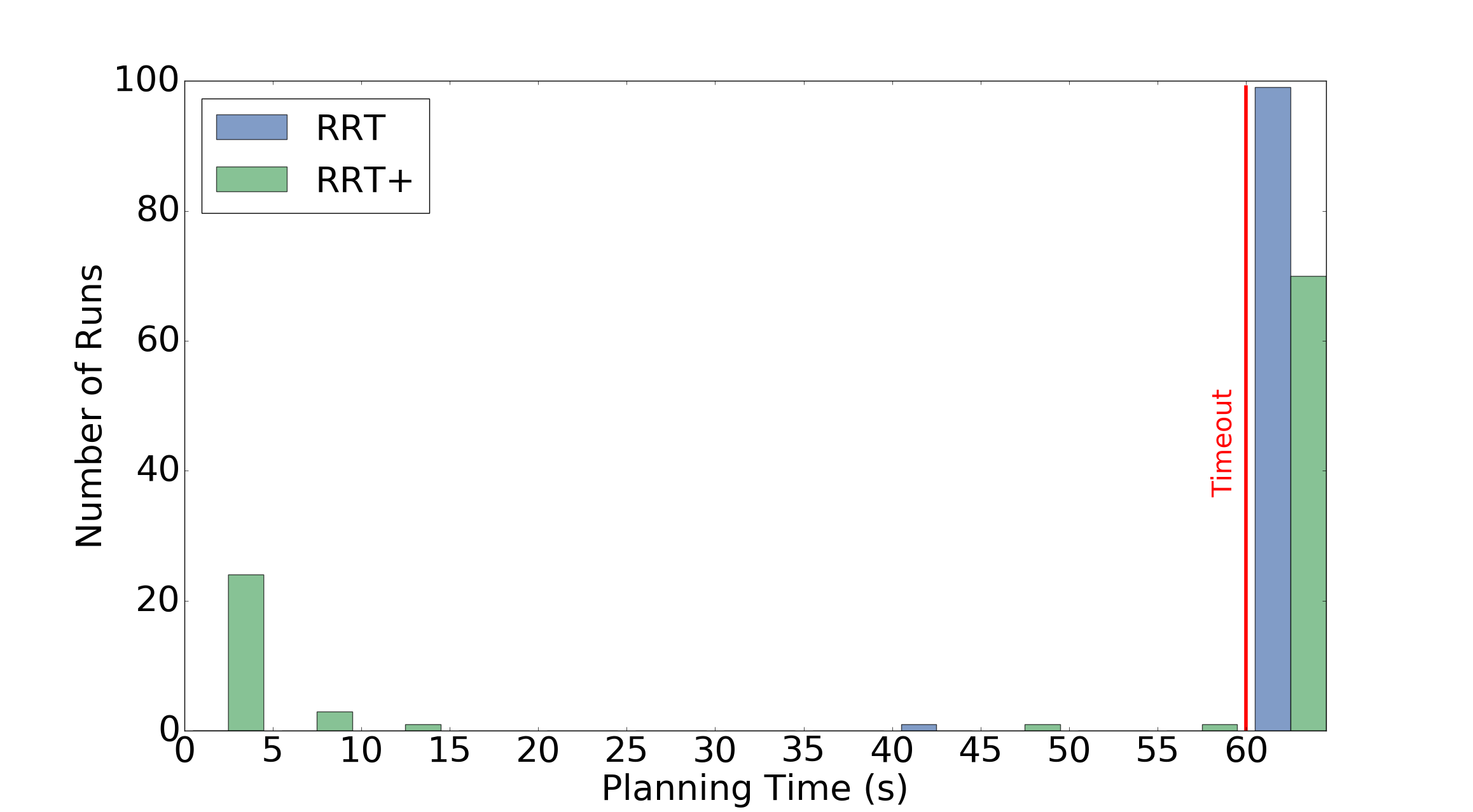}}}\\
     \subfigure[]{\fbox{\includegraphics[width=0.45\textwidth]{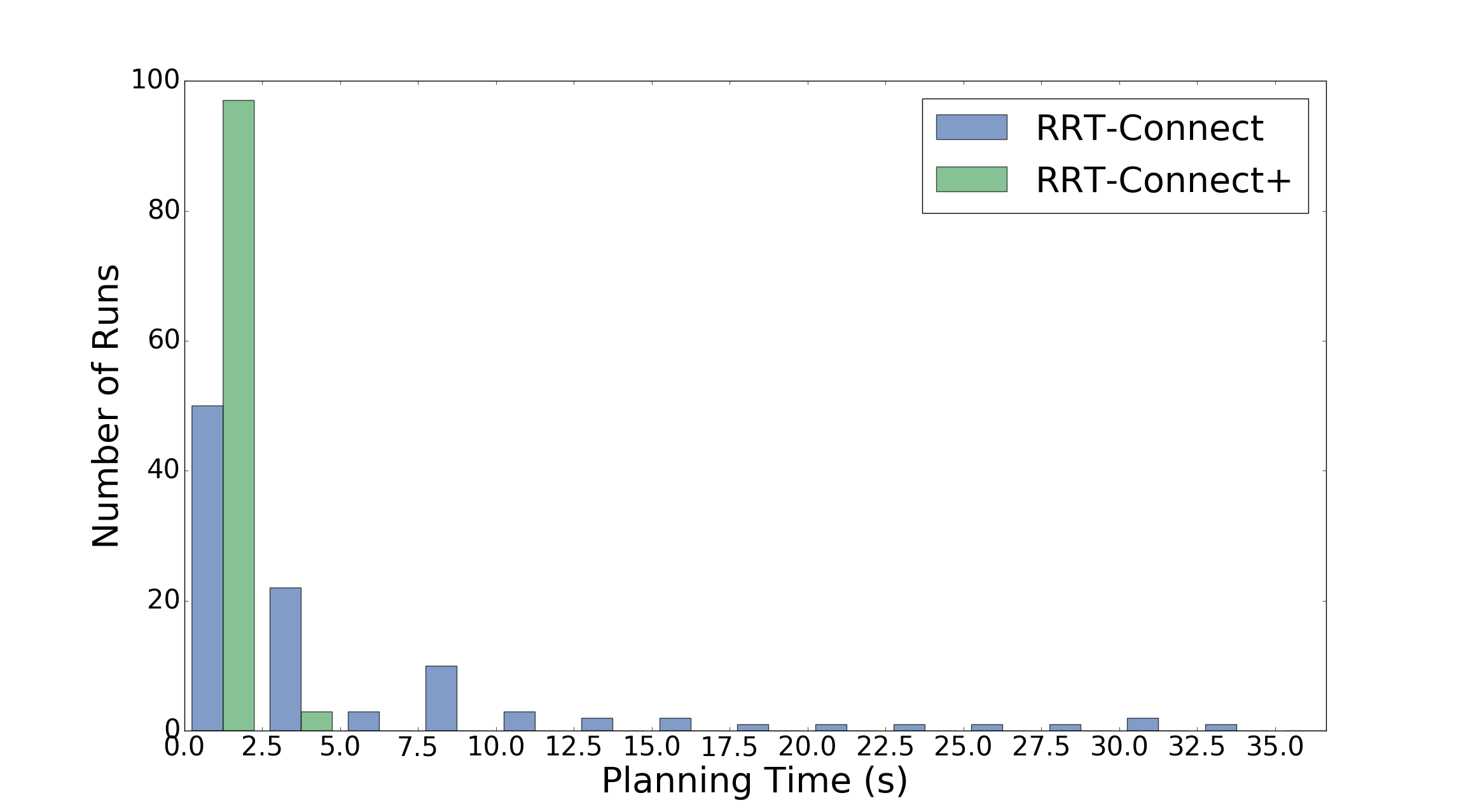}}}\\
     \subfigure[]{\fbox{\includegraphics[width=0.45\textwidth]{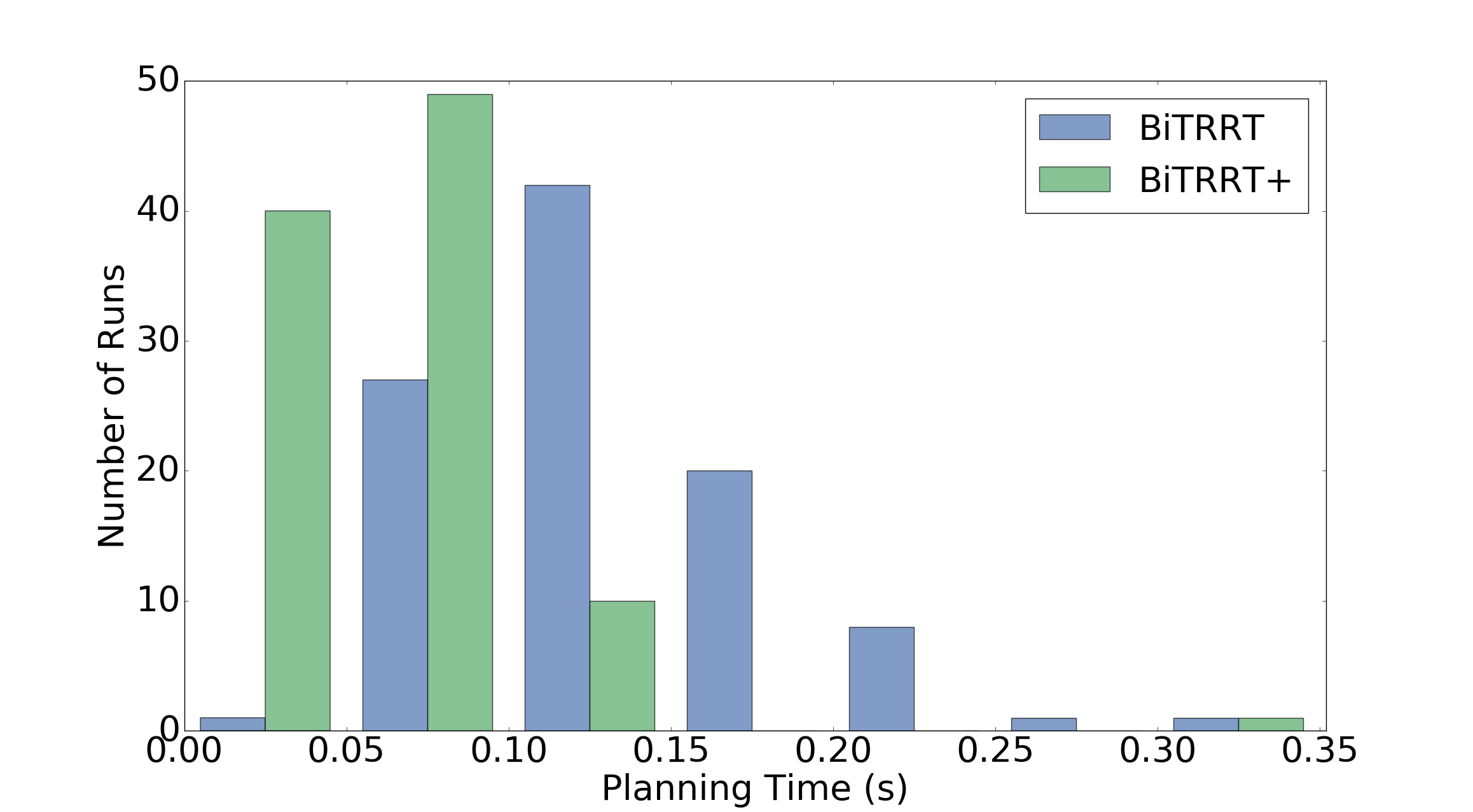}}}\\
     \end{tabular}

\caption{The histograms comparing the enhanced planners, with the original ones. The red line at 60 seconds indicates the Timeout, and the results after that line should be considered failures.}
\label{results}
\end{figure}

\begin{figure}[t]
\centering
\fbox{\includegraphics[width=0.45\textwidth, clip=true]{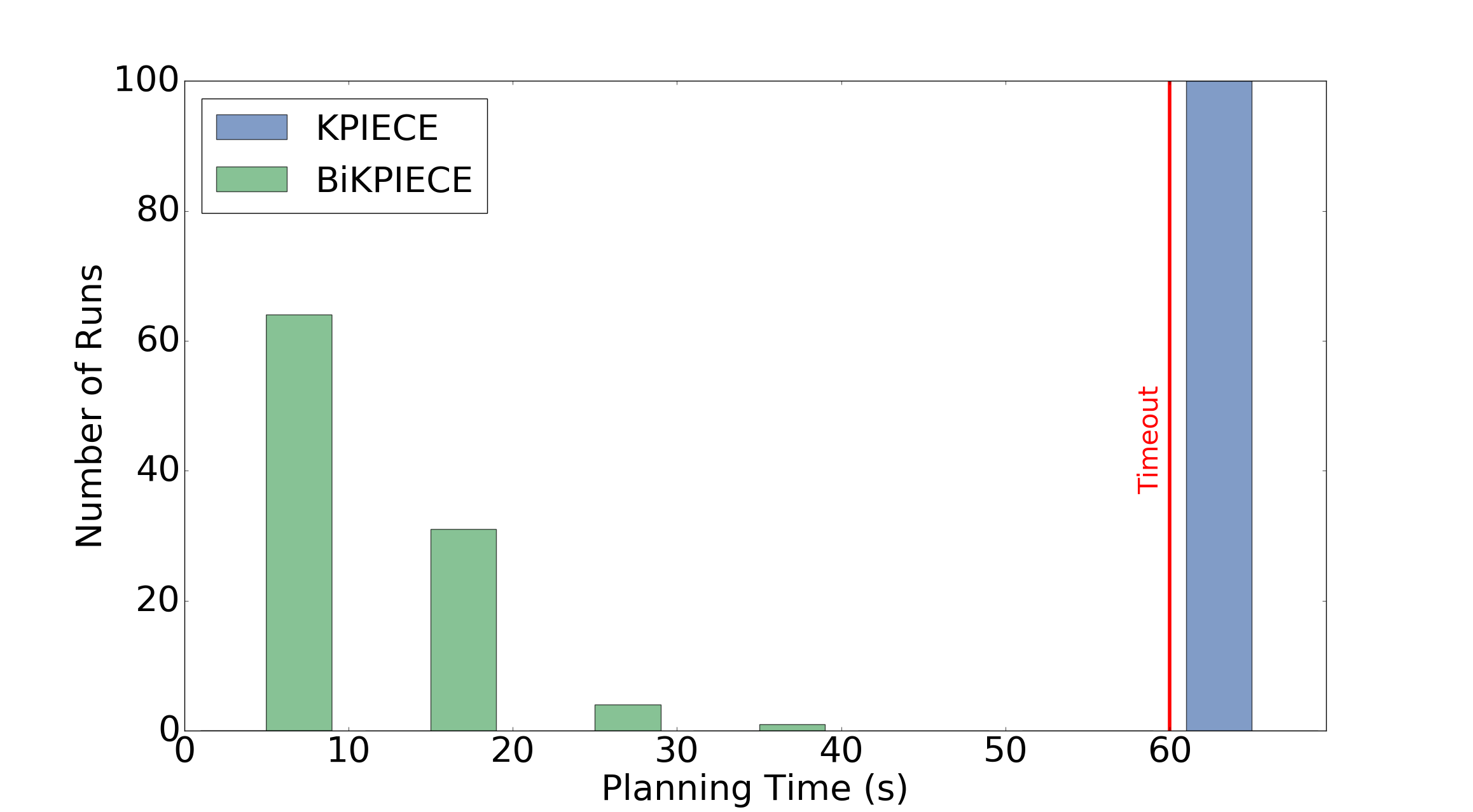}}
\caption{The histograms of KPIECE and BiKPIECE. The red line at 60 seconds indicates the Timeout, and the results after that line should be considered failures.}
\label{fig:kpiece}
\end{figure}

As shown in Figure~\ref{results}, every single enhanced planner, performed much better that the corresponding ones. RRT$^+$-Connect with BiTRRT$^+$ much better than BiKPIECE, and KPIECE, whose results presented in Figure~\ref{fig:kpiece}. Non bidirectional planners had difficulty finding a solution, showing the difficulty of the problem close to the goal region. The enhanced planners, even when they were failing in the case of RRT$^+$, produced solutions very quickly showing the advantage of sampling in lower-dimensional subspaces. Similarly to the experiments presented in the previous section, the fastest planner among all was the BiTRRT$^+$. Dominantly the solutions from RRT$^+$- were found in a 8-dimensional subspace, and from RRT$^+$-Connect and BiTRRT$^+$ in a 6-dimensional. 

\section{Conclusion} 
\label{sec:Conclusion}
We studied a novel method for potentially accelerating motion planning in high dimensional configuration spaces by sampling in subspaces of progressively increasing dimension. The method provides, on average, solutions much faster than the original RRT-based methods.  \invis{The motivation of the work is that in many unpredictable situations that a robotic system can encounter, fast solutions can guarantee its survival while belated optimal solutions may cause severe damages.}

The approach is general enough to be applied to a broad variety of motion
planning problems.  For example, our experiments show potential for planning
with costmaps via adaptation of the bidirectional T-RRT. 
The enhanced planners can also be adjusted to each problem and provide results even faster. The study clearly shows that such methods have great potential in solving very fast, seemingly difficult problems and further investigation is needed.

\invis{Additionally, RRT$^+$ arises new problems that are both related to applications
in complex robotic systems and the further development of the idea.}
Future work will target a number of important questions.  First, it would be
very interesting to find an efficient way to choose subspaces
that are more likely to contain solutions. Second, it is important to provide a
general way to identify when a new iteration should begin, using a metric of the
expansion of the tree, and eliminating the parameters. 
Prior work~\cite{esposito2013conditional} provides an estimation of the coverage in an efficient analytical way. Moreover, by using the previous metric, it is possible to identify when
the tree overcame a difficult area and then reduce the dimensionality of the
search, in order to accelerate the results further. For both of the above problems discrete methods such as the one of {\c{S}}ucan and Kavraki~\cite{alexandru2009performance} should be also considered. 

Currently, there are two planners in OMPL that in some cases outperform the
RRT$^+$ planners in our experiments: KPIECE which uses random 2D or 3D projections to estimate the coverage efficiently and STRIDE which samples non-uniformly with a bias
to narrow spaces. Our study can extend the first planner to sample strictly in subspaces of 
$\C$ and use 2D, 3D or 4D projections to estimate the coverage, and the second planner, can be used with the enhancement to accelerate the solutions in the lower dimensional subspaces where narrow passages are expected to be common. The integration of the ideas underlying the proposed method with those planners to accelerate the results is left for future work.

Lastly, we plan to explore the ability of the method to
efficiently produce paths that satisfy some natural constraints of each system. Since the subspaces are defined only by simple constraints between the DoFs, subspaces that satisfy some dynamic constraints can be explored by defining dynamic constraints between the different DoFs. Similarly, the ability of the method to find alternative solutions by avoiding exploring non-desired areas shows some potential.

\section*{ACKNOWLEDGMENT}

The authors would like to thank the generous support of the Google Faculty Research Award and the National Science Foundation grants (NSF 0953503, 1513203, 1526862, 1637876).

\invis{Here is Jason's grant info. There should be two:

This material is based upon work supported by the National Science
Foundation under Grant No. 1526862.
This material is based upon work supported by the National Science
Foundation under Grant No. 0953503.}



\bibliographystyle{template/IEEEtran}
\bibliography{ref}

\end{document}